\documentclass{article}

\usepackage[preprint,nonatbib]{neurips_2022}

\usepackage[utf8]{inputenc} 
\usepackage[T1]{fontenc}    
\usepackage{url}            
\usepackage{booktabs}       
\usepackage{amsfonts}       
\usepackage{nicefrac}       
\usepackage{microtype}      
\usepackage[dvipsnames,table,xcdraw]{xcolor} 
\usepackage{multirow}

\usepackage{blindtext}
\newcommand{\tableCellHeight}{1}
\newcommand{\tabstyle}[1]{
  \setlength{\tabcolsep}{#1}
  \renewcommand{\arraystretch}{\tableCellHeight}
  \centering
  \small
}

\usepackage{graphicx}
\usepackage{amsmath}
\usepackage{amssymb}
\usepackage{subfiles}

\usepackage{times}
\usepackage{epsfig}

\usepackage[ruled,linesnumbered]{algorithm2e}

\usepackage{float}
\usepackage{rotating}

\usepackage{enumitem}
\usepackage{makecell}
\usepackage{rotating}
\usepackage{mmstyle}
\newcommand{\romannum}[1]{\romannumeral #1} 
\usepackage{subcaption}
\usepackage{color, colortbl}
\definecolor{tabhighlight}{HTML}{e5e5e5}
\definecolor{improvement}{RGB}{225,97,78}
\definecolor{citecolor}{HTML}{0071bc}

\usepackage{marvosym}
\makeatletter
\def\@fnsymbol#1{\ensuremath{\ifcase#1\or \textsuperscript{~\Letter}\or \ddagger\or
   \mathsection\or \mathparagraph\or \|\or **\or \dagger\dagger
   \or \ddagger\ddagger \else\@ctrerr\fi}}
\makeatother

\usepackage[pagebackref,breaklinks,colorlinks,citecolor=citecolor]{hyperref}
\usepackage{threeparttable}
\usepackage{wrapfig}

\title{Neural Prompt Search}

\author{%
  Yuanhan Zhang\quad Kaiyang Zhou\quad Ziwei Liu\thanks{Corresponding author}\\
  S-Lab, Nanyang Technological University, Singapore \\
  \texttt{ \{yuanhan002, kaiyang.zhou, ziwei.liu\}@ntu.edu.sg} \\
}

\begin{document}

\maketitle
\begin{abstract}
The size of vision models has grown exponentially over the last few years, especially after the emergence of Vision Transformer. This has motivated the development of parameter-efficient tuning methods, such as learning adapter layers or visual prompt tokens, which allow a tiny portion of model parameters to be trained whereas the vast majority obtained from pre-training are frozen. However, designing a proper tuning method is non-trivial: one might need to try out a lengthy list of design choices, not to mention that each downstream dataset often requires custom designs. In this paper, we view the existing parameter-efficient tuning methods as ``prompt modules'' and propose \textbf{Neural prOmpt seArcH (NOAH)}, a novel approach that learns, for large vision models, the optimal design of prompt modules through a neural architecture search algorithm, specifically for each downstream dataset. By conducting extensive experiments on over 20 vision datasets, we demonstrate that NOAH (\romannum{1}) is superior to individual prompt modules, (\romannum{2}) has good few-shot learning ability, and (\romannum{3}) is domain-generalizable. 
The code and models are available at \url{https://github.com/Davidzhangyuanhan/NOAH}.
\end{abstract}
\section{Introduction}

\begin{figure}[t]
\centering
\includegraphics[width=0.99\textwidth]{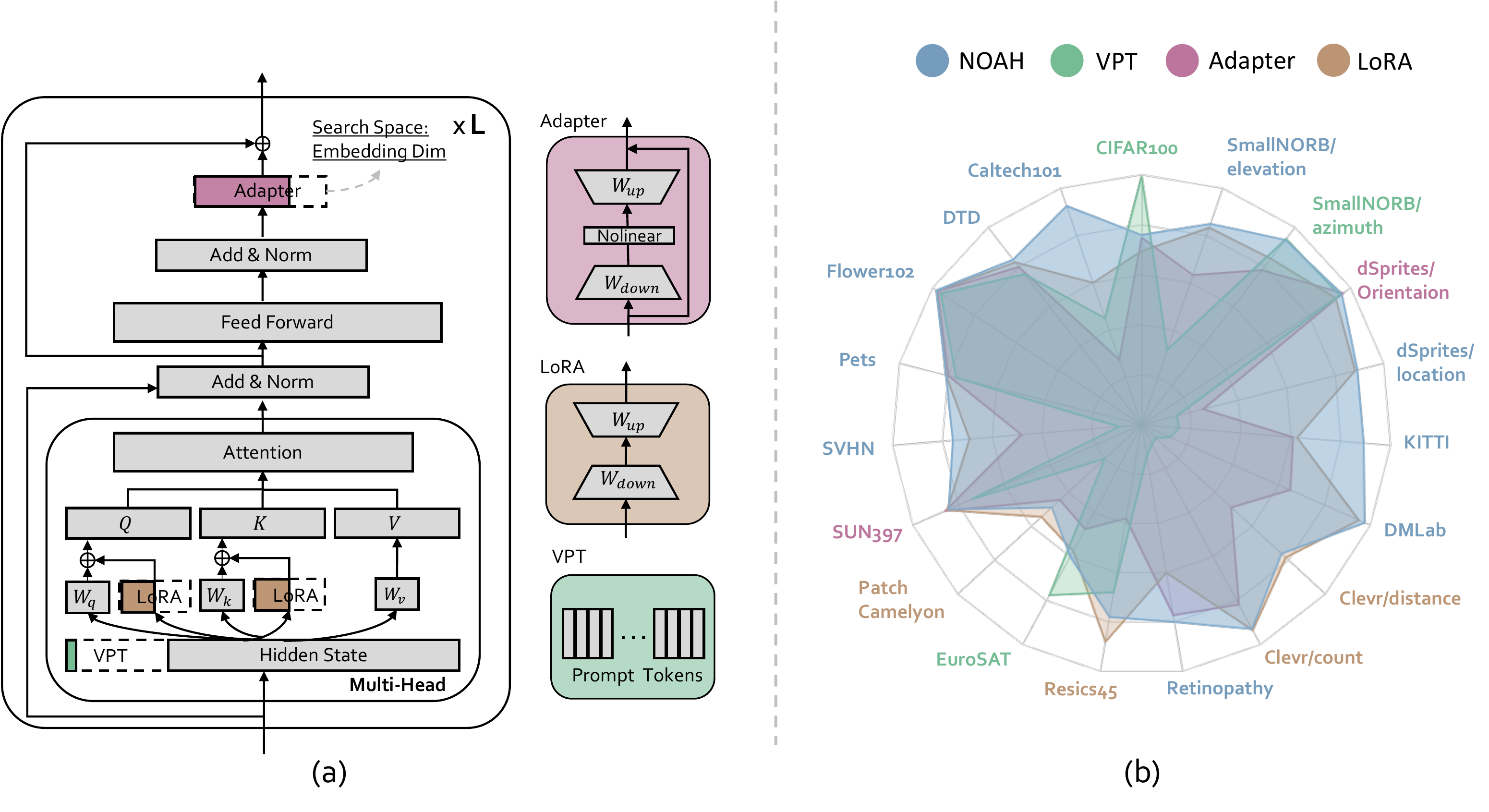}
\caption{
Our approach, neural prompt search, or NOAH for short, subsumes three representative parameter-efficient tuning methods (i.e., Adapter~\cite{houlsby2019parameter}, LoRA~\cite{hu2021lora} and VPT~\cite{jia2022visual}) and learns from data the optimal design through neural architecture search (a). The approach is motivated by the observation that none of the three individuals shows dominance on the VTAB-1k benchmark (b). The colors of the datasets' names indicate which method performs the best. Clearly, NOAH is the best overall approach.
}
\label{fig:motivation}
\end{figure}

The size of vision models has grown exponentially from tens of millions a few years ago (\eg, ResNet~\cite{he2016deep}) to today's hundreds of millions~\cite{dosovitskiy2020image}, or even billions~\cite{brown2020language,devlin2018bert,reed2022generalist}, for Transformers~\cite{vaswani2017attention}. Such an increase can cause a number of problems to transfer learning~\cite{houlsby2019parameter,jia2022visual}, and the first and foremost is that fine-tuning becomes more difficult as large model size can easily lead to overfitting in a typical-sized dataset, let alone the increase of compute and storage costs.

Recently, there is a growing interest in developing parameter-efficient tuning methods~\cite{houlsby2019parameter,hu2021lora,jia2022visual}. The key idea is to insert a tiny trainable module to a large pre-trained model and only adjust its parameters by optimizing some task-specific losses like the cross-entropy for classification problems. The most representative methods are Adapter~\cite{houlsby2019parameter}, Low-Rank Adaptation (LoRA)~\cite{hu2021lora}, and Visual Prompt Tuning (VPT)~\cite{jia2022visual}. As exemplified in Fig.~\ref{fig:motivation}(a), Adapter is a bottleneck-shaped neural network appended to a network block's output; LoRA is a ``residual'' layer consisting of rank decomposition matrices; VPT prepends additional tokens to the input of a Transformer block, which can be seen as adding learnable ``pixels.''

By evaluating the three parameter-efficient tuning methods on a commonly-used transfer learning benchmark, \ie, VTAB-1k~\cite{zhai2019large}, we identify a couple of critical issues. \emph{First}, none of the three methods performs consistently well on all datasets, as illustrated in Fig.~\ref{fig:motivation}(b). For instance, when it comes to scene structure understanding tasks, VPT outperforms Adapter and LoRA on SmallNORB/
azimuth~\cite{lecun2004learning}, but its performance plunges on SmallNORB/elevation~\cite{lecun2004learning} and Clevr/count~\cite{johnson2017clevr}, which is largely behind the two competitors. The results suggest that, for a specific dataset, one needs to perform an extensive evaluation on different tuning methods in order to identify the most suitable one. \emph{Second}, performance is found to be sensitive to the selection of model parameters, such as Adapter's feature dimension or the token length in VPT---this is also observed by Jia et al.~\cite{jia2022visual} that the optimal token length in VPT varies from 1 to 200 on different datasets.

In this work, we view the existing parameter-efficient tuning methods as \emph{prompt modules} and propose to \emph{automatically search for the optimal prompt design from data via a neural architecture search (NAS) algorithm}. Specifically, we introduce the concept of \underline{N}eural pr\underline{O}mpt se\underline{A}rc\underline{H} (NOAH) for large vision models, particularly those equipped with the Transformer block~\cite{dosovitskiy2020image}. The search space is constructed by subsuming Adapter~\cite{houlsby2019parameter}, LoRA~\cite{hu2021lora} and VPT~\cite{jia2022visual} into each Transformer block, as depicted in Fig.~\ref{fig:motivation}(a). The specific model parameters, including the feature dimension for Adapter and LoRA and the token length for VPT, are determined by a one-shot NAS algorithm.

We conduct extensive experiments on VTAB-1k~\cite{zhai2019large}, which is composed of 19 diverse vision datasets and covers a wide spectrum of visual domains like objects, scenes, textures and satellite imagery. The results show that NOAH significantly outperforms the individual prompt modules on 10 out of 19 datasets while the performance on the remaining is highly competitive (see Fig.~\ref{fig:motivation}(b) for an overview of the results). We also evaluate on few-shot learning and domain generalization where the results also confirm the superiority of NOAH to the hand-crafted prompt modules.

Our contributions are summarized as follows. (\romannum{1}) We present a systematic study of three representative prompt modules and expose some critical issues associated with performance and efficiency. (\romannum{2}) A novel concept, neural prompt search, is proposed to address the challenge of hand-engineering prompt modules. (\romannum{3}) An efficient NAS-based implementation of NOAH is provided. (\romannum{4}) We demonstrate that NOAH is better than individual prompt modules in downstream transfer learning, few-shot learning, and domain generalization. The models and code will be released.

\section{Neural Prompt Search}
\label{sec:method}

\subsection{Background}
\label{sec:method;subsec:bg}

\paragraph{Vision Transformer}
We first briefly review Vision Transformer (ViT)~\cite{dosovitskiy2020image}, to which our approach is mainly applied. ViT consists of alternating blocks of multihead self-attention (MSA) and multi-layer perceptron (MLP). Given an input sequence $x \in \mathbb{R}^{N \times D}$ where $N$ denotes the token length and $D$ is the embedding dimension, MSA first maps $x$ to queries $Q \in \mathbb{R}^{N \times d}$, keys $K \in \mathbb{R}^{N \times d}$ and values $V \in \mathbb{R}^{N \times d}$ using three projection matrices, $W_q \in \mathbb{R}^{D \times d}$, $W_k \in \mathbb{R}^{D \times d}$ and $W_v \in \mathbb{R}^{D \times d}$, respectively, where $d$ means the hidden dimension. Then, MSA computes the weighted sums over the values based on the self-attention between the queries and keys,
\begin{equation}\label{eq:msa}
\operatorname{Attention}(Q, K, V) = \operatorname{softmax}(\frac{QK^T}{\sqrt{d}})V,
\end{equation}
where $\frac{1}{\sqrt{d}}$ is a scaling factor.

Below we briefly review the three representative---and top-performing---parameter-efficient tuning methods, \ie, Adapter~\cite{houlsby2019parameter}, LoRA~\cite{hu2021lora} and VPT~\cite{jia2022visual}, which will be incorporated into our search space. An illustration of these three methods can be found in Fig.~\ref{fig:motivation}(a). Note that VPT has been studied for vision models while Adapter and LoRA have only been studied for language models.

\paragraph{Adapter}
is essentially a bottleneck-like neural network consisting of a down-sample layer $W_{down} \in \mathbb{R}^{d\times r}$ and an up-sample layer $W_{up} \in \mathbb{R}^{r\times d}$, where $r$ denotes the down-sampled dimension. A non-linear activation function $\phi(\cdot)$, such as ReLU, is inserted in-between. The computation can be formulated as
\begin{equation}
\label{eqn:adapter}
    h' =  \phi(h W_{down}) W_{up},
\end{equation}
where $h \in \mathbb{R}^{N \times d}$ is a normalized output of the MLP in a Transformer block.

\paragraph{LoRA}
aims to update the two projection layers, $W_q$ (for queries) and $W_k$ (for keys), in an indirect way by optimizing their rank-decomposed changes, $\bigtriangleup W_q = W_{q}^{down}W_{q}^{up}$ and $\bigtriangleup W_k = W_{k}^{down}W_{k}^{up}$, where $W_{q/k}^{down} \in \mathbb{R}^{D \times r}$ and $W_{q/k}^{up} \in \mathbb{R}^{r \times d}$ ($r$ is the down-projection dimension). For a specific input $x$, we have
\begin{equation}
\label{eqn:lora}
    Q =  x W_q + s \cdot x W_{q}^{down} W_{q}^{up}, \quad
    K =  x W_k + s \cdot x W_{k}^{down} W_{k}^{up},
\end{equation}
where $s$ is a fixed scaling parameter for modulating the updates.

\paragraph{Visual Prompt Tuning (VPT)}
prepends a set of learnable tokens to the input of a Transformer block, which can be viewed as adding some learnable pixels in the input space. We investigate the best-performing version, VPT-Deep, which applies prompt tuning to multiple layers~\cite{jia2022visual}. We call this module VPT for brevity hereafter and formulate it in mathematical terms below. A typical input $x \in \mathbb{R}^{N \times D}$ to a Transformer block contains a learnable class token [CLS] of $D$-dimension and a sequence of image patch embeddings $\text{E} = \{ \text{e}_i | \text{e}_i \in \mathbb{R}^D, i = 1, ..., N-1 \}$ where the positional embeddings are omitted. VPT adds $m$ learnable tokens, $\text{P} = \{ \text{p}_k | \text{p}_k \in \mathbb{R}^D, k = 1, ..., m \}$, to $x$, which then becomes
\begin{equation}
\label{eqn:vpt}
x = [ \text{CLS}, \text{P}, \text{E} ].
\end{equation}

\subsection{Prompt Search Algorithm}
\label{sec:method;subsec:prompt_search_alg}

As discussed, none of the individual parameter-efficient tuning methods, or \emph{prompt modules} called in this paper, shows dominance in the transfer learning benchmark. Our approach, neural prompt search (NOAH), incorporates Adapter~\cite{houlsby2019parameter}, LoRA~\cite{hu2021lora} and VPT~\cite{jia2022visual} into each Transformer block and learns the design that best suits a dataset through neural architecture search (NAS). Specifically, we employ a one-shot NAS algorithm, AutoFormer~\cite{chen2021autoformer}, for prompt module search. Our supernet is a ViT-like model composed of 12 Transformer blocks (layers). Below we detail the search space and how the search is done.

\paragraph{Search Space}
As shown in Fig.~\ref{fig:motivation}(a), we embed the three prompt modules into each Transformer block following the guidelines proposed in the original work~\cite{houlsby2019parameter,hu2021lora,jia2022visual}. Concretely, we install VPT in the input position, add LoRA alongside the two projection matrices as residuals, and insert Adapter after the normalized output of the MLP. The search space mainly contains the model parameters associated with the three prompt modules. Specifically, each prompt module has two sets of parameters to search from: (\romannum{1}) the embedding dimension $\in \{5, 10, 50, 100\}$ or $\{1, 5, 10\}$; (\romannum{2}) the depth $\in \{3, 6, 9, 12\}$. A depth means up to which layer a module is applied, \eg, $\text{depth} = 3$ for VPT means layers 0, 1 and 2 have VPT installed while the remaining layers, 3 to 11, do not have VPT.\footnote{Zero-based indexing is adopted.} For VPT, the embedding dimension means the token length whereas for Adapter and LoRA, the embedding dimension means the down-sampled dimension, \ie, $r$.

\paragraph{Supernet Training}
The supernet, as mentioned, has 12 Transformer layers, each containing the three prompt modules with full embedding dimension, \ie, 100. During each forward pass, a subnet is randomly sampled from the supernet for training. Specifically, for \emph{each} prompt module, a depth is first sampled from $\{3, 6, 9, 12\}$ to determine which layers should have the module. Then, for \emph{each} layer within the depth range, an embedding dimension is chosen from $\{5, 10, 50, 100\}$ or $\{1, 5, 10\}$, all with a uniform probability. Note that only the prompt modules' parameters are learned while the pre-trained model is kept fixed. AutoFormer~\cite{chen2021autoformer} allows the weights in each prompt module to be entangled during training, meaning that different weights are maximally shared, \eg, in a VPT module, if 100 tokens are selected for training, the previously trained tokens, such as 50, will be reused and trained together with other 50 tokens. This way, as suggested in AutoFormer~\cite{chen2021autoformer}, leads to faster convergence and low memory cost.

\begin{figure}[t]
\centering
\includegraphics[width=\linewidth]{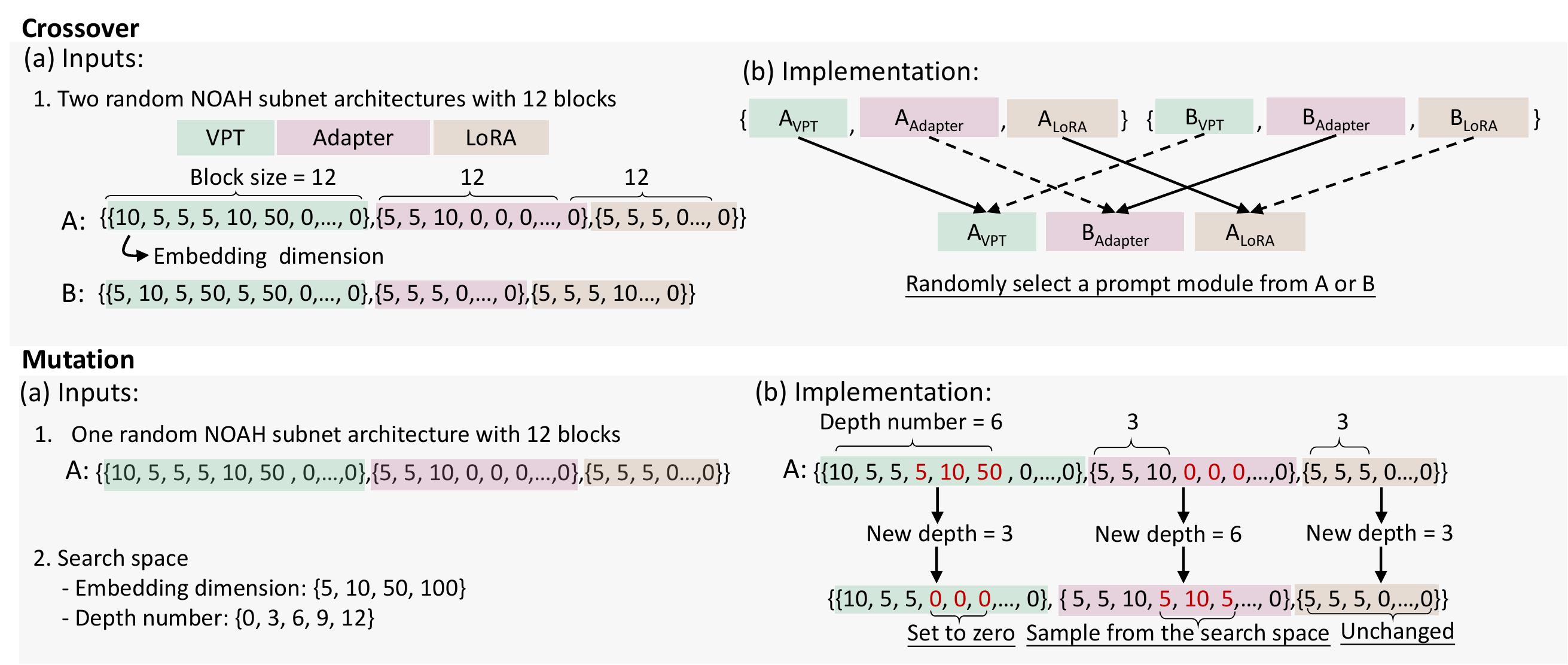}
\caption{Illustration of crossover and mutation in the evolutionary search method.  
}
\label{fig:evolution_search}
\end{figure}

\paragraph{Evolutionary Search}
After the supernet is trained, evolutionary search is conducted to obtain the optimal subnet architecture under a parameter size limit~\cite{chen2021autoformer}. Specifically, we first select $K$ random architectures, from which the top $k$ architectures (with the best performance) are used as parents to produce the next generation through crossover and mutation. For crossover, two candidates are randomly chosen and crossed to produce a ``child'' architecture. For mutation, a candidate mutates its prompt module design with a probability. See Fig.~\ref{fig:evolution_search} for an illustration.

\section{Experiments}
\label{sec:exp}

In this section, we mainly address the following questions: (\romannum{1}) Is NOAH better than the individual prompt modules? (\romannum{2}) Can NOAH work in a few-shot setting? (\romannum{3}) Are models learned by NOAH robust to domain shift? The answers are discussed in Sec.~\ref{sec:exp;subsec:vtab}, \ref{sec:exp;subsec:fewshot} and \ref{sec:exp;subsec:dg}, respectively. We also conduct some analyses in Sec.~\ref{sec:exp;subsec:analysis} to have a deeper understanding of NOAH, such as what a subnet looks like and whether it is transferable beyond the dataset in which the architecture was found.

\paragraph{Baselines}
The main competitors are the three representative prompt modules subsumed by NOAH, which are Adapter~\cite{houlsby2019parameter}, LoRA~\cite{hu2021lora} and VPT~\cite{jia2022visual}. Among them, only VPT is specifically designed for vision models while the other two are originally developed for language models. We also compare two common fine-tuning methods on the VTAB-1k benchmark: full tuning (Full) and linear probing (Linear). Full simply tunes the entire model parameters whereas Linear freezes the pre-trained part and only adjusts the newly added linear classification layer.\footnote{Note that all methods have a new linear classification layer to learn.} It is worth mentioning that Full has been considered as a strong baseline in existing studies~\cite{jia2022visual,houlsby2019parameter}.

\paragraph{Implementation Details}
We keep the training parameters identical across all experiments throughout this paper. ViT-B/16~\cite{dosovitskiy2020image} pre-trained on ImageNet-22K~\cite{deng2009imagenet} is used as the base model, which is strong enough so the results are fair and convincing. The supernet for NOAH is trained for 500 epochs and the ultimate subnet is trained for 100 epochs--- note that ``subnet'' means the prompt modules/architectures. Since the AutoFormer~\cite{chen2021autoformer} algorithm allows a subnet to be used without retraining, we demonstrate later that the subnet found by NOAH without retraining is also comparable to the retrained one. The evolutionary search in NOAH takes 5 epochs in total and each step of random pick/crossover/mutation produces 50 new subnets. The probability for crossover and mutation is set to 0.2, which follows AutoFormer~\cite{chen2021autoformer}. The individual prompt modules, \ie, Adapter~\cite{houlsby2019parameter}, LoRA~\cite{hu2021lora} and VPT~\cite{jia2022visual}, are constructed using the best recipes suggested by the original papers (also trained for 100 epochs; see the Supplementary for more details). The parameter sizes for Adapter, LoRA and VPT are 0.16M, 0.29M and 0.64M, respectively. For fair comparison, we set the upper-limit of parameter size of the final subnet in NOAH to 0.64M so the resulting size would be comparable to the baselines. More implementation details including image augmentation and other hyper-parameters are provided in the Supplementary.

\subsection{Experiments on VTAB-1k}
\label{sec:exp;subsec:vtab}

\paragraph{Datasets}
We choose the VTAB-1k~\cite{zhai2019large} benchmark to evaluate the transfer learning performance of our approach. VTAB-1k consists of 19 vision datasets, which are clustered into three groups: Natural, Specialized and Structured. The Natural group contains natural images that are captured by standard cameras and cover a broad spectrum of concepts including generic, fine-grained and abstract objects. The Specialized group contains images captured by specialist equipment for remote sensing (like aerial images) and medical purposes. The Structured group is designed specifically for scene structure understanding, such as object counting, depth prediction and orientation prediction. Each dataset in VTAB-1k contains 1,000 labeled examples, which are split into a \texttt{train} (80\%) and a \texttt{val} (20\%) set (the latter is used for hyper-parameter tuning), while the test data comes from the original test set. The final model used for evaluation is trained using the full 1,000 examples in each dataset. Top-1 classification accuracy is used as the performance measure.

\begin{table}[t]
\tabstyle{2pt}
\caption{
\textbf{Full results on the VTAB-1k benchmark}. The first block contains conventional tuning methods while the second block contains parameter-efficient tuning methods, which is the main focus in this paper. NOAH achieves the best overall performance, which is 1\% higher on average than the individual prompt modules.
}
\label{tab:main_results}
\scriptsize
    \begin{tabular}{lc | ccccccc | cccc | cccccccc | c}
    \toprule
    & & \multicolumn{7}{c|}{\textbf{Natural}} & \multicolumn{4}{c|}{\textbf{Specialized}} & \multicolumn{8}{c|}{\textbf{Structured}} \\
    & \rotatebox{90}{\# param (M)} & \rotatebox{90}{Cifar100} & \rotatebox{90}{Caltech101} & \rotatebox{90}{DTD} & \rotatebox{90}{Flower102} & \rotatebox{90}{Pets} & \rotatebox{90}{SVHN}  & \rotatebox{90}{Sun397} & \rotatebox{90}{Camelyon} & \rotatebox{90}{EuroSAT}   & \rotatebox{90}{Resisc45}  & \rotatebox{90}{Retinopathy} & \rotatebox{90}{Clevr-Count} & \rotatebox{90}{Clevr-Dist}  & \rotatebox{90}{DMLab} & \rotatebox{90}{KITTI-Dist}  & \rotatebox{90}{dSpr-Loc} & \rotatebox{90}{dSpr-Ori}   & \rotatebox{90}{sNORB-Azim}  & \rotatebox{90}{sNORB-Ele} & \rotatebox{90}{Average}   \\
    \midrule
    Full~\cite{jia2022visual} & 85.8  & \textbf{68.9} & \textbf{87.7} & \textbf{64.3} & 87.2  & \textbf{86.9} & \textbf{87.4} & 38.8 & \textbf{79.7}          & \textbf{95.7}         & \textbf{84.2}          & 73.9           & \textbf{56.3}          & \textbf{58.6}          & \textbf{41.7} & \textbf{65.5} & \textbf{57.5}          & \textbf{46.7}          & \textbf{25.7}           & \textbf{29.1}           & \textbf{68.9}  \\ 
    Linear~\cite{jia2022visual} & 0.04  & 64.4          & 85.0          & 63.2          & \textbf{97.0}          & 86.3          & 36.6          & \textbf{51.0}                   & 78.5          & 87.5          & 68.5          & \textbf{74.0}          & 34.3          & 30.6          & 33.2          & 55.4 & 12.5          & 20.0          & 9.6           & 19.2          & 57.6  \\
    \midrule
VPT~\cite{jia2022visual} & 0.64   & \textbf{78.8} & 90.8 & 65.8 & 98.0 & 88.3 & 78.1 & 49.6 & 81.8 & \textbf{96.1} & 83.4 & 68.4 & 68.5 & 60.0 & 46.5 & 72.8 & 73.6 & 47.9 & \textbf{32.9} & 37.8 &  72.0 \\
Adapter~\cite{houlsby2019parameter} & 0.16 & 69.2 & 90.1 & 68.0 & 98.8 & 89.9 & 82.8 & \textbf{54.3} & 84.0 & 94.9 & 81.9 & 75.5 & 80.9 & 65.3 & 48.6 & 78.3 & 74.8 & \textbf{48.5} & 29.9 & 41.6 &  73.9 \\
LoRA~\cite{hu2021lora}& 0.29 & 67.1 & 91.4 & 69.4 & 98.8 & \textbf{90.4} & 85.3 & 54.0 & \textbf{84.9} & 95.3 & \textbf{84.4} & 73.6 & \textbf{82.9} & \textbf{69.2} & 49.8 & 78.5 & 75.7 & 47.1 & 31.0 & 44.0 &  74.5 \\
\rowcolor{tabhighlight} NOAH & 0.43 & 69.6 & \textbf{92.7} & \textbf{70.2} & \textbf{99.1} & \textbf{90.4} & \textbf{86.1} & 53.7 & 84.4 & 95.4 & 83.9 & \textbf{75.8} & 82.8 & 68.9 & \textbf{49.9} & \textbf{81.7} & \textbf{81.8} & 48.3 & 32.8 & \textbf{44.2} &  \textbf{75.5} \\\bottomrule
    \end{tabular}
\end{table}


\begin{figure}[t]
    \centering
    \includegraphics[width=\textwidth]{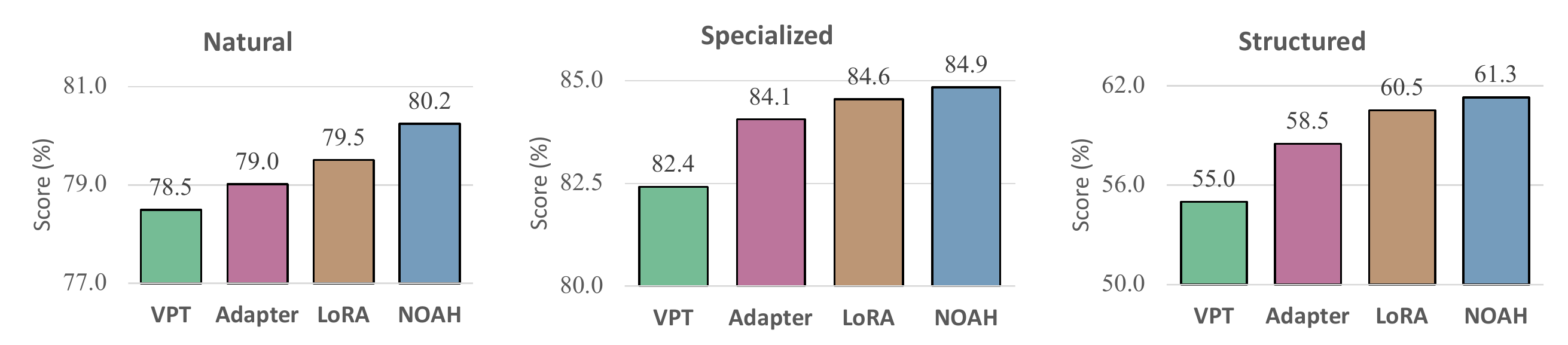}
    \caption{\textbf{Group-wise average results on VTAB-1k}. NOAH performs the best in the Natural and Structured groups while its performance in the Specialized group is similar to that of LoRA---but NOAH does not require a manual search over the architecture and hyper-parameters.}
    \label{fig:vtab_avg}
\end{figure}

\paragraph{Results}
Table~\ref{tab:main_results} presents the full results on the VTAB-1k benchmark. A high-level summary is shown earlier in Fig.~\ref{fig:motivation}(b). The average performance within each group is summarized in Fig.~\ref{fig:vtab_avg}. We have the following observations.

\emph{Observation 1: Overall, NOAH is the best parameter-efficient tuning method.}
First and foremost, we demonstrate that searching for the optimal combination of the individual prompt modules works the best. This is evidenced by the 1\% average gain over the strongest prompt module, \ie, LoRA. Given the diversity of the benchmark, the 1\% average gain can be considered to be significant. It is also worth mentioning that Adapter was previously proved to be the best-performing prompt module in NLP~\cite{mao2021unipelt}, but in our study for computer vision tasks, LoRA takes over the seat. This further confirms that search is a better option than hand-engineering in practice.

\emph{Observation 2: NOAH slightly dims in the Specialized group.}
The results suggest that NOAH's weakness seems to be in the Specialized tasks where the individual modules achieve the on-par performance: NOAH's results are not too far from those of the competitors, \eg, NOAH's 84.9 vs LoRA's 84.6 on average. And while NOAH is superior on a Retinopathy, it lags on the other datasets, especially on the EuroSAT. Since the individual modules require a manual search over architecture and hyper-parameters, NOAH is more compelling.

\subsection{Experiments on Few-Shot Learning}
\label{sec:exp;subsec:fewshot}

\begin{figure}[t]
\centering
\includegraphics[width=\linewidth]{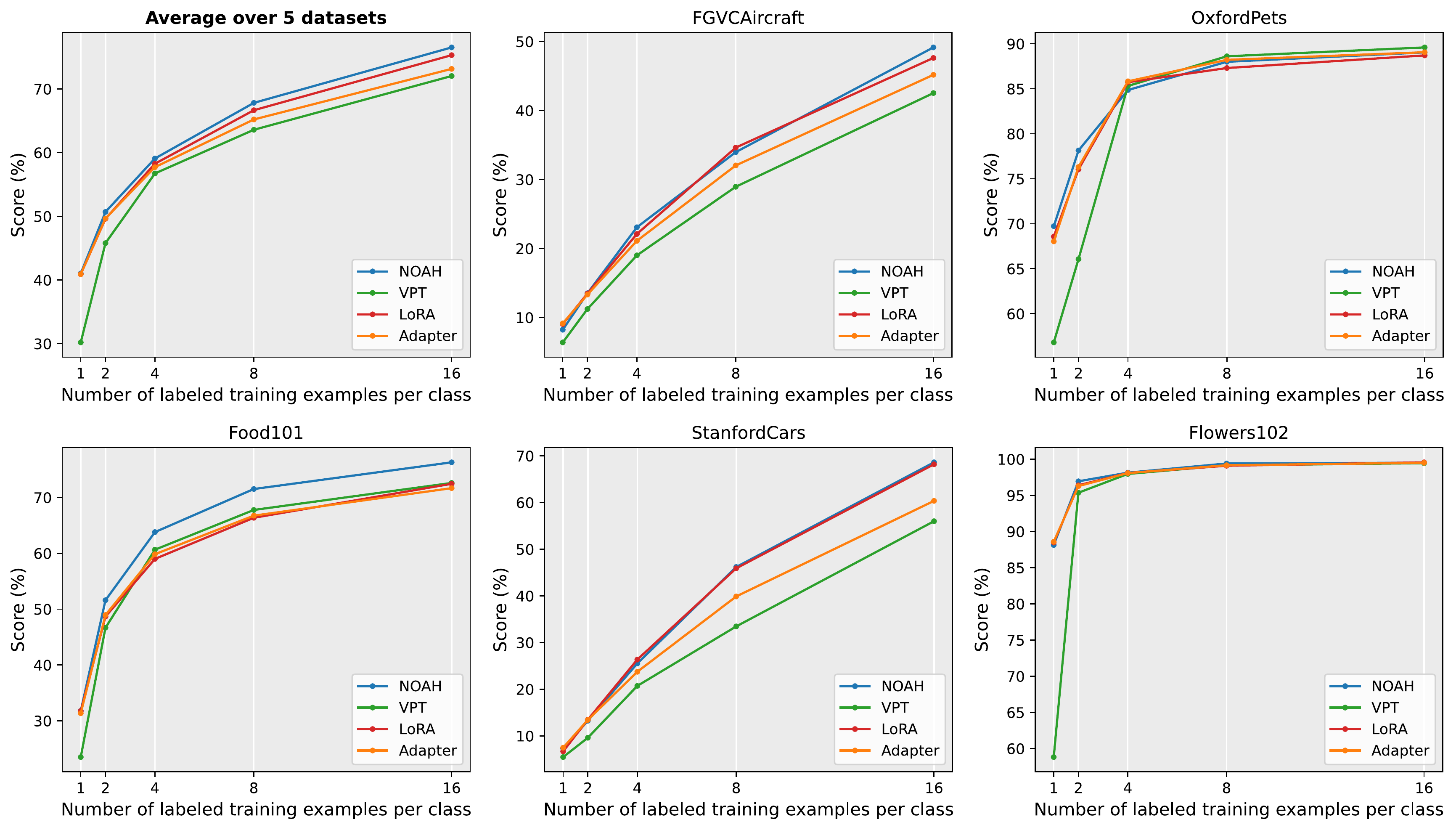}
\caption{
\textbf{Results of few-shot learning on five fine-grained visual recognition datasets}. NOAH beats the individual modules on average.
}
\label{fig:few_shot}
\end{figure}

\paragraph{Datasets}
We choose five fine-grained visual recognition datasets, which include Food101~\cite{bossard2014food}, OxfordFlowers102~\cite{nilsback2006visual}, StandfordCars~\cite{krause20133d}, OxfordPets~\cite{parkhi2012cats}, and FGVCAircraft~\cite{maji2013fine}. The categories in these datasets cover a wide range of visual concepts closely related to our daily life: food, plant, vehicle and animal. We follow existing studies~\cite{zhou2021learning,radford2021learning} to evaluate on 1, 2, 4, 8 and 16 shots, which are sufficient for observing the trend.

\paragraph{Results}
The results are summarized in Fig.~\ref{fig:few_shot}. In terms of the average performance, we can observe that: (\romannum{1}) In the low-data regime like 1 or 2 shots, NOAH, LoRA and Adapter perform similarly but VPT largely lags behind; (\romannum{2}) NOAH shows clear dominance when more shots become available, \eg, with 16 shots the gap between NOAH and the runner-up is around 2\%. By looking at the individual graphs, we can see that none of the individual prompt modules performs consistently well on all datasets, which, again, justifies that search is better than hand-engineering.

\subsection{Experiments on Domain Generalization}
\label{sec:exp;subsec:dg}

\begin{wraptable}{r}{0.55\textwidth}
    \tabstyle{5pt}
    \caption{\textbf{Results on domain generalization}. NOAH is significantly better than the individual prompt modules on the four domain-shifted datasets.}
    \label{tab:dg}
    \begin{tabular}{l ccccc}
    \toprule
    & \textbf{Source} & \multicolumn{4}{c}{\textbf{Target}} \\ \cmidrule(lr){2-2} \cmidrule(lr){3-6}
    & ImageNet & -V2 & -Sketch & -A & -R \\
    \midrule
    Adapter~\cite{houlsby2019parameter} &70.5 & 59.1 & 16.4 & 5.5 & 22.1 \\
    VPT~\cite{jia2022visual} & 70.5 & 58.0 & 18.3 & 4.6 & 23.2 \\
    LoRA~\cite{hu2021lora} & 70.8 & 59.3 & 20.0 & 6.9 & 23.3 \\
    \rowcolor{tabhighlight} NOAH & \textbf{71.5}  & \textbf{66.1}  & \textbf{24.8}  & \textbf{11.9}  & \textbf{28.5}  \\
    \bottomrule
    \end{tabular}
\end{wraptable}

\paragraph{Datasets}
Since domain shift is ubiquitous in real-world applications~\cite{zhou2021domain}, we are interested to know how our search-based approach compares with the individual prompt modules in terms of domain generalization ability. Following prior studies~\cite{zhou2021learning,zhou2022conditional}, we first train a model on ImageNet~\cite{deng2009imagenet} (using 16 shots per category) and then directly test it on four other variants of ImageNet that undergo different types of domain shift. Specifically, the test datasets include (\romannum{1}) ImageNetV2~\cite{recht2019imagenet}, which is collected from different sources than ImageNet but following the same collection protocol, (\romannum{2}) ImageNet-Sketch~\cite{wang2019learning}, which is composed of sketch images of the same 1,000 classes in ImageNet, (\romannum{3}) ImageNet-A~\cite{hendrycks2021natural}, which contains adversarially-filtered images, (\romannum{4}) ImageNet-R~\cite{hendrycks2021many}, which is a rendition of ImageNet. Both ImageNet-A and -R have 200 classes derived from a subset of ImageNet's 1000 classes. All results are averaged over three random seeds.

\paragraph{Results}
Table~\ref{tab:dg} compares NOAH with the three individual prompt modules. On ImageNet, which is the source dataset, the gap between NOAH and the individual modules is small, which is about 1\%. However, on the four test datasets, NOAH demonstrates significantly stronger robustness than the baselines: over 6.8\%, 4.8\%, 5\% and 5.2\% improvements on -V2, -Sketch, -A and -R, respectively. The results, together with those from previous subsections, justify that our search-based approach is superior to the individual prompt modules.


\begin{figure}[t]
\centering
\includegraphics[width=\linewidth]{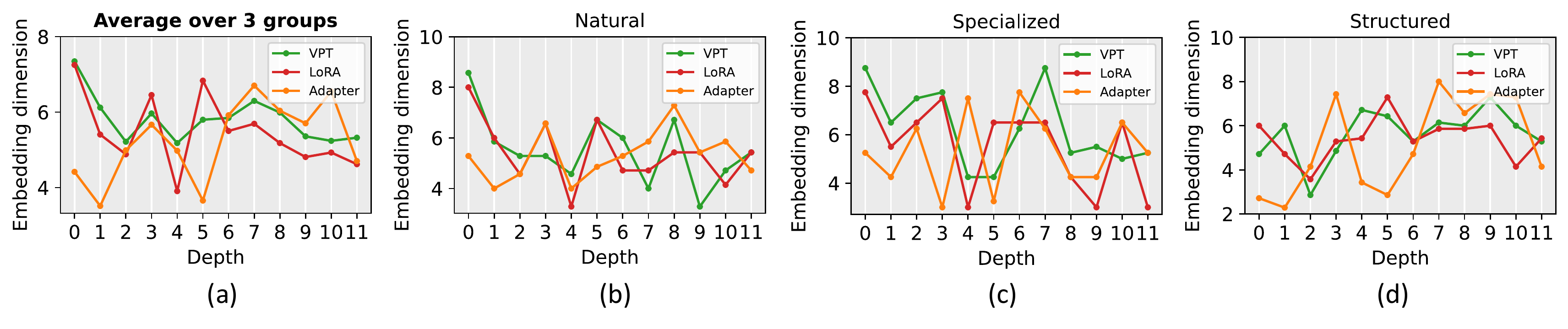}
\caption{
\textbf{Average subnets (architectures) for the three groups in VTAB-1k}. Adapter and LoRA tend to live in deep layers while VPT is found nearly in all depths. The demands for VPT (indicated by the embedding dimension) differ in different groups. The co-existence of the three modules, especially in deep layers, serves as strong evidence of their complementarity, and such a synergy is difficult to obtain by hand-engineering.
}
\label{fig:architechture}
\end{figure}

\subsection{Further Analysis}
\label{sec:exp;subsec:analysis}

\paragraph{Architecture of Subnet}
A key question to answer is: how does NOAH's subnet, \ie, the ultimate architecture, look like. To make the results convincing, we visualize the average architecture---instead of individual ones---found within each group of VTAB-1k, as well as the global average over all datasets, in Fig.~\ref{fig:architechture}. The x-axis represents the network depth while the y-axis represents the embedding dimension. An intriguing observation is that Adapter and LoRA, \emph{across all groups} (Fig.~\ref{fig:architechture}(a)), mainly appear in deep layers with the embedding dimension larger than four and reduced to zero in some shallow layers. In contrast, VPT can be found nearly in all depths (layers), but the dimensions vary significantly in different groups, which indicates different demands for VPT. For instance, in the Natural group (Fig.~\ref{fig:architechture}(b)), shallow layers need more VPT modules; but in the Structured group (Fig.~\ref{fig:architechture}(d)), deep layers need more VPT modules. Moreover, the co-existence of the three modules, especially in deep layers, suggests that they are complementary to each other---\emph{such a synergy is difficult to obtain by manual design}. In summary, the observed high variances in the module designs strongly indicate that search is much more efficient than hand-engineering when it comes to developing parameter-efficient tuning methods.

\begin{figure}[t]
  \centering
  \includegraphics[width=\linewidth]{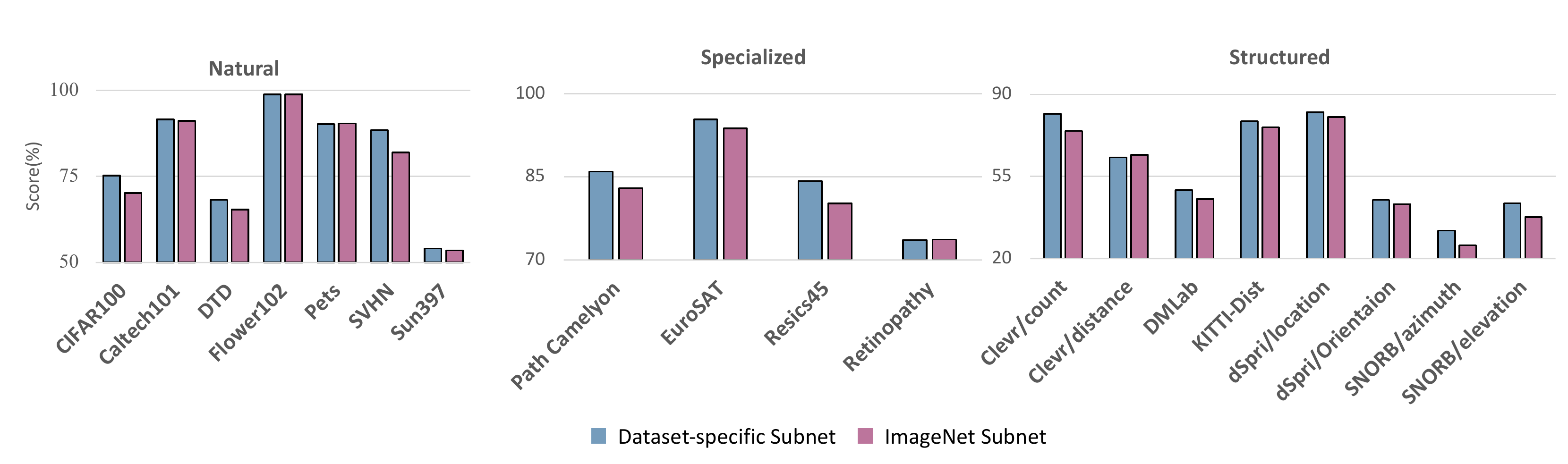}
  \caption{\textbf{Evaluation on the transferability of subnets}. Dataset-specific subnet means the architecture is found from the target dataset. ImageNet subnet means the architecture is found from ImageNet and transferred to the target dataset. All target datasets come from VTAB-1k. In general, better transferability is achieved when the source and target datasets are closer, and vice versa.
  }
  \label{fig:transferability}
\end{figure}

\paragraph{Transferability of Subnet}
As discussed previously, the subnet (\ie, architecture) found for different datasets differs dramatically. Here we study whether, or in what circumstances, the subnet found from one dataset can be transferred to another. To this end, we train NOAH on ImageNet and apply the ultimate subnet to the VTAB-1k benchmark where the model is retrained and evaluated. To measure transferability, we compare the ImageNet subnet with the dataset-specific subnets on VTAB-1k. Fig.~\ref{fig:transferability} shows the comparisons. Overall, the gap between the ImageNet subnet and the 19 dataset-specific subnets on VTAB-1k is below 3\%, meaning that NOAH has fair transferability. By digging deeper into the results, we find that the transfer gap is smaller when the source (\ie, ImageNet) and target datasets are closer, and vice versa. For instance, the gaps in the Natural group are less than 1\%, which make sense because the ImageNet images and those from the Natural group share similar visual concepts, such as generic objects, flowers and animals.

\begin{wraptable}{r}{0.55\textwidth}
    \tabstyle{7pt}
    \caption{
    \textbf{With vs without retraining NOAH's subnets}. The results show that there is no significant difference between them, suggesting that retraining can be safely removed in practice if the compute resource is limited.
    }
    \label{tab:retrain_vs_inherit}
    \begin{tabular}{l cccc}
    \toprule
    & Nat. & Spe. & Str. & Average \\
    \midrule
    VPT~\cite{jia2022visual} & 78.5  & 82.4 & 55.0  & 72.0   \\
    Adapter~\cite{houlsby2019parameter} & 79.0  &  84.1 & 58.5 & 73.9 \\
    LoRA~\cite{hu2021lora} & 79.5 & 84.6  & 60.5  & 74.5   \\
    \rowcolor{tabhighlight} NOAH (\emph{inherited}) & 79.5 & 84.2  & 58.7 & 74.1 \\ 
    \rowcolor{tabhighlight} NOAH (\emph{retrained}) & \textbf{80.2} & \textbf{84.9} & \textbf{61.3} & \textbf{75.5} \\
    \bottomrule
    \end{tabular}
\end{wraptable}

\paragraph{With vs Without Retraining}
Thanks to the weight entanglement strategy in AutoFormer~\cite{chen2021autoformer}, the subnet extracted from the supernet can be directly deployed for use without retraining. To verify if such a rule also applies to NOAH, we compare the subnets with and without retraining on VTAB-1k. The results averaged over each group are shown in Table~\ref{tab:retrain_vs_inherit} where we observe that NOAH without retraining (denoted as \emph{inherited}) is still competitive: the inherited version still outperforms the VPT and Adapter. The results suggest that the retraining cost can be safely removed without incurring any significant loss.
\section{Related Work}


\subsection{Parameter-Efficient Tuning}
A recent trend in transfer learning is to develop parameter-efficient tuning methods~\cite{jia2022visual,houlsby2019parameter,hu2021lora,he2021towards,zhong2021factual,zhou2021learning}, which is spurred by the rapid increase in model size. Existing methods can be generally divided into two groups. The first group fine-tunes a small portion of the internal parameters, such as biases~\cite{zaken2021bitfit}. The second group adds tiny learnable modules like Adapter~\cite{houlsby2019parameter} or LoRA~\cite{hu2021lora}, which is more relevant to our research and thus the focus here. Adapter~\cite{houlsby2019parameter} and LoRA~\cite{hu2021lora} essentially share similar architectures---both look like a bottleneck---but are installed at different places: Adapter is often installed at the output of a block while LoRA is treated as residuals to the projection matrices in a Transformer~\cite{vaswani2017attention} block. It is worth noting that these methods are first studied in natural language processing (NLP) since pre-trained language models~\cite{devlin2018bert,brown2020language} typically have an enormous parameter size that reaches the billion level. Another popular design in NLP is prompt learning~\cite{zhong2021factual,li2021prefix,lester2021power}, which turns some text prompt tokens into learnable vectors. Such an idea has recently been applied to vision-language models~\cite{zhou2021learning,zhou2022conditional,lu2022prompt} and is also the source of inspiration for the recently proposed VPT~\cite{jia2022visual}, which adds learnable ``pixels'' to the input of ViT~\cite{dosovitskiy2020image}.

More relevant to our work are those trying to unify different parameter-efficient tuning methods~\cite{he2021towards,mao2021unipelt}. He \etal~\cite{he2021towards} build a connection between Adapter and prompt learning and cast the problem into the learning of a modification vector, which leads to a unified view and a stronger baseline. UNIPELT~\cite{mao2021unipelt} is another unified framework, which subsumes several prompt modules in a block and learns a set of gating functions to selectively activate them. Our work differs from these studies in two crucial ways: (\romannum{1}) we target \emph{computer vision} problems whereas the previous studies~\cite{he2021towards,mao2021unipelt} focus on NLP; (\romannum{2}) we unify prompt modules from the NAS perspective with a much more \emph{fine-grained} control over the model hyper-parameters (\eg, token length and embedding dimension). This allows our model to be deployed in a resource-constrained environment. In the future, we plan to apply our approach to NLP.

\subsection{Neural Architecture Search}
Neural architecture search (NAS) consists of two crucial components: search space and search algorithm. A search space can subsume various designs of how neurons are connected~\cite{zoph2018learning}, diverse combinations of model hyper-parameters~\cite{zoph2016neural}, or different arrangements of specific modules like normalization layers~\cite{zhou2021osnet}. When it comes to the search algorithm part, the community has witnessed significant advances: from costly methods like reinforcement learning~\cite{zoph2016neural} or evolutionary search~\cite{real2017large} to more efficient ones based on weight-sharing~\cite{pham2018efficient} or differentiable optimization~\cite{liu2018darts}. The most relevant work to ours is AutoFormer~\cite{chen2021autoformer}, which is a one-shot NAS method focusing on Transformer~\cite{vaswani2017attention} models. AutoFormer features a weight entanglement strategy, which allows different subnets sampled from a big supernet to share weights among each other. Our work leverages AutoFormer to solve the problem of engineering parameter-efficient tuning methods, which we hope can inspire future work to address efficient transfer learning.
\section{Discussion, Limitation and Future Work}
With the proliferation of large-scale pre-training data~\cite{zhang2022bamboo,radford2021learning,yuan2021florence}, model size in neural networks has also been increased correspondingly in order to reach a certain learning capacity. On the other hand, the rapid increase in model size has also spurred interests in developing efficient transfer learning methods~\cite{jia2022visual,zhou2021learning}.

Our research presents timely studies on how some recently-proposed parameter-efficient tuning methods, or prompt modules, fare in computer vision problems. Crucially, our studies expose a critical issue that, for any specific downstream dataset, hand-designing an optimal prompt module is extremely challenging. More importantly, we for the first time solve the problem from a NAS perspective and demonstrate the potential of our search-based approach in terms of downstream transfer learning performance, the ability to work in low-data regimes, and robustness to domain shift, which is ubiquitous in real-world data~\cite{zhou2021domain}.

Our studies also unveil some intriguing phenomena. In particular, we find that the ultimate subnet exhibits different architectural patterns for the three prompt modules across datasets of different natures. Since neural networks' features, as often suggested~\cite{zeiler2014visualizing}, progress from low-level visual primitives in bottom layers to high-level abstractions in top layers, the aforementioned findings entail that different prompt modules work best for features at different levels. We hope such findings and insights can inspire future work on designing more advanced prompt modules.

In terms of limitations, NOAH requires additional training for the supernet, which inevitably increases the development cost. Moreover, as suggested by the few-shot learning results, NOAH's advantages become clearer when more labeled images are available. In other words, NOAH would require more labels to unleash its full power in practice. For future work, we plan to dig deeper into the mechanisms behind NOAH for better interpretation of the intriguing results and apply NOAH to broader application domains beyond computer vision, such as NLP.
\appendix
\section{Implementation Details}
\subsection{Datasets}
Table~\ref{tab:dataset_statistics} briefly introduces dataset that we used. 

\begin{table}[h]
\tabstyle{2pt}
\caption{The first block introduces the information of the datasets in the VTAB-1k~\cite{zhai2019large} benchmark. The datasets in the second block are five fine-grain dataset used in the CoOp~\cite{zhou2021learning} few-shot learning benchmark. Especially, \emph{Research only} indicates that this dataset is for research purposes only.
}
\label{tab:dataset_statistics}
\begin{tabular}{lllllll}
\toprule
                          & Dataset              & \#Classes & Train                    & Val   & Test   & License         \\ \midrule
\multirow{19}{*}{VTAB-1k~\cite{zhai2019large}} & CIFAR100~\cite{krizhevsky2009learning}             & 100       & \multirow{19}{*}{800/1,000}                 & \multirow{19}{*}{200}   & 10,000 & Research only         \\
                          & Caltech101~\cite{fei2004learning}           & 102       &                 &    & 6,084  & Research only         \\
                          & DTD~\cite{cimpoi14describing}                 & 47        &                 &    & 1,880  & Research only         \\
                          & Oxford-Flowers102~\cite{nilsback2006visual}    & 102       &                 &    & 6,149  & Research only         \\
                          & Oxford-Pets~\cite{parkhi2012cats}          & 37        &                 &    & 3,669  & CC BY-SA 4.0    \\
                          & SVHN~\cite{netzer2011reading}                 & 10        &                 &    & 26,032 & CC              \\
                          & Sun397~\cite{xiao2010sun}               & 397       &                 &    & 21,750 & Research only         \\
                          & Patch Camelyon~\cite{Veeling2018qh}       & 2         &                 &    & 32,768 & CC0             \\
                          & EuroSAT~\cite{helber2019eurosat}              & 10        &                 &    & 5,400  & Research only         \\
                          & Resisc45~\cite{cheng2017remote}             & 45        &                 &    & 6,300  & Research only         \\
                          & Retinopathy~\cite{kaggle2015retinopathy}         & 5         &                 &    & 42,670 & Research only         \\
                          & Clevr/count~\cite{johnson2017clevr}          & 8         &                 &    & 15,000 & CC BY 4.0       \\
                          & Clevr/distance~\cite{johnson2017clevr}       & 6         &                 &    & 15,000 & CC BY 4.0       \\
                          & DMLab~\cite{beattie2016deepmind}                & 6         &                 &    & 22,735 & Research only         \\
                          & KITTI-Dist~\cite{geiger2013vision}           & 4         &                 &    & 711    & CC BY-NC-SA 3.0 \\
                          & dSprites/location~\cite{matthey2017dsprites}    & 16        &                 &    & 73,728 & Research only         \\
                          & dSprites/orientation~\cite{matthey2017dsprites} & 16        &                 &    & 73,728 & Research only         \\
                          & SmallNORB/azimuth~\cite{lecun2004learning}    & 18        &                 &    & 12,150 & Research only   \\
                          & SmallNORB/elevation~\cite{lecun2004learning}  & 18         &                 &    & 12,150 & Research only   \\ \midrule
\multirow{5}{*}{Few-shot~\cite{zhou2021learning}} & Food-101~\cite{bossard2014food}             & 101       & \multirow{5}{*}{(1/2/4/8/16)*(\#Classes)} & 20,200 & 30,300  & Research only         \\
                          & Stanford Cars~\cite{krause20133d}        & 196       &  & 1,635  & 8,041   & Research only   \\
                          & Oxford-Flowers102~\cite{nilsback2006visual}    & 102       &  & 1,633  & 2,463   & Research only         \\
                          & FGVC-Aircraft~\cite{maji2013fine}        & 100       &  & 3,333  & 3,333   & Research only   \\
                          & Oxford-Pets~\cite{parkhi2012cats}          & 37        &  & 736   & 3,669   & CC BY-SA 4.0  \\\bottomrule
\end{tabular}
\end{table}

\subsection{Training Details}
\paragraph{Augmentation}
For the VTAB-1k~\cite{zhai2019large}, we follows its default augmentation settings, implementing the resizing and normalization for input images. Specifically, we resize a input image to  $224 \times 224$, followed by normalizing it with ImageNet~\cite{deng2009imagenet} means and standard deviation. For few-shot learning and domain generalization experiments, we implement color-jitters with the factor as 0.4, and RandAugmentation with magnitude equals 9, magnitude standard deviation equals 0.5.
\paragraph{Hyperparameters}
We consistently set the embedding dimension of Adapter~\cite{houlsby2019parameter} and LoRA~\cite{hu2021lora} as 8. We set prompt length of VPT~\cite{jia2022visual} following the paper. For few-shot learning (FS) and domain generalization (DG) experiments, we consistently set the VPT prompt length as 8. We use AdamW optimizer~\cite{loshchilov2017decoupled} with the cosine scheduler. The weight decay equals $1$e$-3$, warm-up epochs equals 10, and the batch size equals 64. Other hyperparameters are shown in below.


\begin{table}[h]
\tabstyle{2pt}
\caption{\textbf{Hyperparameters}. Others includes few-shot learning and domain generalization.
}
\begin{tabular}{lcccccl}
\toprule
    & \multicolumn{2}{c}{Learning Rate} & \multicolumn{2}{c}{Dimension}  & \multirow{2}{*}{Depth}   \\ 
    & VTAB & Others                      & VTAB & Others &   \\ \midrule
VPT &   1$e$-3 &   5$e$-3 &  \multirow{5}{*}{\{1,5,10\}}  & \multirow{5}{*}{\{5,10,50,100\}}   & \multirow{5}{*}{\{3,6,9,12\}}\\
Adapter       & 1$e$-3 &   5$e$-3   &  &    \\
LoRA          &  1$e$-3 &   5$e$-3 &   &      \\
NOAH-supernet &  5$e$-4 &   5$e$-4 &  &    \\
NOAH-subnet   &  1$e$-3  &   5$e$-3  &   \\\bottomrule
\end{tabular}
\end{table}




\clearpage
{
\bibliographystyle{plain}
\bibliography{main.bbl}

\begin{thebibliography}{10}

\bibitem{beattie2016deepmind}
Charles Beattie, Joel~Z Leibo, Denis Teplyashin, Tom Ward, Marcus Wainwright,
  Heinrich K{\"u}ttler, Andrew Lefrancq, Simon Green, V{\'\i}ctor Vald{\'e}s,
  Amir Sadik, et~al.
\newblock Deepmind lab.
\newblock {\em arXiv preprint arXiv:1612.03801}, 2016.

\bibitem{bossard2014food}
Lukas Bossard, Matthieu Guillaumin, and Luc~Van Gool.
\newblock Food-101--mining discriminative components with random forests.
\newblock In {\em European conference on computer vision (ECCV)}, pages
  446--461. Springer, 2014.

\bibitem{brown2020language}
Tom Brown, Benjamin Mann, Nick Ryder, Melanie Subbiah, Jared~D Kaplan, Prafulla
  Dhariwal, Arvind Neelakantan, Pranav Shyam, Girish Sastry, Amanda Askell,
  et~al.
\newblock Language models are few-shot learners.
\newblock {\em Advances in neural information processing systems (NeuIPS)},
  33:1877--1901, 2020.

\bibitem{chen2021autoformer}
Minghao Chen, Houwen Peng, Jianlong Fu, and Haibin Ling.
\newblock Autoformer: Searching transformers for visual recognition.
\newblock In {\em Proceedings of the IEEE/CVF International Conference on
  Computer Vision (ICCV)}, pages 12270--12280, 2021.

\bibitem{cheng2017remote}
Gong Cheng, Junwei Han, and Xiaoqiang Lu.
\newblock Remote sensing image scene classification: Benchmark and state of the
  art.
\newblock {\em Proceedings of the IEEE}, 105(10):1865--1883, 2017.

\bibitem{cimpoi14describing}
M.~Cimpoi, S.~Maji, I.~Kokkinos, S.~Mohamed, , and A.~Vedaldi.
\newblock Describing textures in the wild.
\newblock In {\em Proceedings of the {IEEE} Conf. on Computer Vision and
  Pattern Recognition ({CVPR})}, 2014.

\bibitem{deng2009imagenet}
Jia Deng, Wei Dong, Richard Socher, Li-Jia Li, Kai Li, and Li~Fei-Fei.
\newblock Imagenet: A large-scale hierarchical image database.
\newblock In {\em Proceedings of the IEEE/CVF Conference on Computer Vision and
  Pattern Recognition (CVPR)}, pages 248--255. Ieee, 2009.

\bibitem{devlin2018bert}
Jacob Devlin, Ming-Wei Chang, Kenton Lee, and Kristina Toutanova.
\newblock Bert: Pre-training of deep bidirectional transformers for language
  understanding.
\newblock {\em arXiv preprint arXiv:1810.04805}, 2018.

\bibitem{dosovitskiy2020image}
Alexey Dosovitskiy, Lucas Beyer, Alexander Kolesnikov, Dirk Weissenborn,
  Xiaohua Zhai, Thomas Unterthiner, Mostafa Dehghani, Matthias Minderer, Georg
  Heigold, Sylvain Gelly, et~al.
\newblock An image is worth 16x16 words: Transformers for image recognition at
  scale.
\newblock {\em arXiv preprint arXiv:2010.11929}, 2020.

\bibitem{fei2004learning}
Li~Fei-Fei, Rob Fergus, and Pietro Perona.
\newblock Learning generative visual models from few training examples: An
  incremental bayesian approach tested on 101 object categories.
\newblock In {\em conference on computer vision and pattern recognition
  workshop}, pages 178--178. IEEE, 2004.

\bibitem{geiger2013vision}
Andreas Geiger, Philip Lenz, Christoph Stiller, and Raquel Urtasun.
\newblock Vision meets robotics: The kitti dataset.
\newblock {\em The International Journal of Robotics Research},
  32(11):1231--1237, 2013.

\bibitem{he2021towards}
Junxian He, Chunting Zhou, Xuezhe Ma, Taylor Berg-Kirkpatrick, and Graham
  Neubig.
\newblock Towards a unified view of parameter-efficient transfer learning.
\newblock {\em arXiv preprint arXiv:2110.04366}, 2021.

\bibitem{he2016deep}
Kaiming He, Xiangyu Zhang, Shaoqing Ren, and Jian Sun.
\newblock Deep residual learning for image recognition.
\newblock In {\em Proceedings of the IEEE/CVF Conference on Computer Vision and
  Pattern Recognition (CVPR)}, pages 770--778, 2016.

\bibitem{helber2019eurosat}
Patrick Helber, Benjamin Bischke, Andreas Dengel, and Damian Borth.
\newblock Eurosat: A novel dataset and deep learning benchmark for land use and
  land cover classification.
\newblock {\em IEEE Journal of Selected Topics in Applied Earth Observations
  and Remote Sensing}, 2019.

\bibitem{hendrycks2021many}
Dan Hendrycks, Steven Basart, Norman Mu, Saurav Kadavath, Frank Wang, Evan
  Dorundo, Rahul Desai, Tyler Zhu, Samyak Parajuli, Mike Guo, et~al.
\newblock The many faces of robustness: A critical analysis of
  out-of-distribution generalization.
\newblock In {\em Proceedings of the IEEE/CVF International Conference on
  Computer Vision (ICCV)}, pages 8340--8349, 2021.

\bibitem{hendrycks2021natural}
Dan Hendrycks, Kevin Zhao, Steven Basart, Jacob Steinhardt, and Dawn Song.
\newblock Natural adversarial examples.
\newblock In {\em Proceedings of the IEEE/CVF Conference on Computer Vision and
  Pattern Recognition (CVPR)}, pages 15262--15271, 2021.

\bibitem{houlsby2019parameter}
Neil Houlsby, Andrei Giurgiu, Stanislaw Jastrzebski, Bruna Morrone, Quentin
  De~Laroussilhe, Andrea Gesmundo, Mona Attariyan, and Sylvain Gelly.
\newblock Parameter-efficient transfer learning for nlp.
\newblock In {\em International Conference on Machine Learning (ICML)}, pages
  2790--2799. PMLR, 2019.

\bibitem{hu2021lora}
Edward~J Hu, Yelong Shen, Phillip Wallis, Zeyuan Allen-Zhu, Yuanzhi Li, Shean
  Wang, Lu~Wang, and Weizhu Chen.
\newblock Lora: Low-rank adaptation of large language models.
\newblock {\em arXiv preprint arXiv:2106.09685}, 2021.

\bibitem{jia2022visual}
Menglin Jia, Luming Tang, Bor-Chun Chen, Claire Cardie, Serge Belongie, Bharath
  Hariharan, and Ser-Nam Lim.
\newblock Visual prompt tuning.
\newblock {\em arXiv preprint arXiv:2203.12119}, 2022.

\bibitem{johnson2017clevr}
Justin Johnson, Bharath Hariharan, Laurens Van Der~Maaten, Li~Fei-Fei,
  C~Lawrence~Zitnick, and Ross Girshick.
\newblock Clevr: A diagnostic dataset for compositional language and elementary
  visual reasoning.
\newblock In {\em Proceedings of the IEEE/CVF Conference on Computer Vision and
  Pattern Recognition (CVPR)}, pages 2901--2910, 2017.

\bibitem{kaggle2015retinopathy}
Kaggle and EyePacs.
\newblock Kaggle diabetic retinopathy detection.
\newblock 2015.

\bibitem{krause20133d}
Jonathan Krause, Michael Stark, Jia Deng, and Li~Fei-Fei.
\newblock 3d object representations for fine-grained categorization.
\newblock In {\em Proceedings of the IEEE international conference on computer
  vision workshops}, pages 554--561, 2013.

\bibitem{krizhevsky2009learning}
Alex Krizhevsky, Geoffrey Hinton, et~al.
\newblock Learning multiple layers of features from tiny images.
\newblock 2009.

\bibitem{lecun2004learning}
Yann LeCun, Fu~Jie Huang, and Leon Bottou.
\newblock Learning methods for generic object recognition with invariance to
  pose and lighting.
\newblock In {\em Proceedings of the IEEE/CVF Conference on Computer Vision and
  Pattern Recognition (CVPR)}, volume~2, pages II--104. IEEE, 2004.

\bibitem{lester2021power}
Brian Lester, Rami Al-Rfou, and Noah Constant.
\newblock The power of scale for parameter-efficient prompt tuning.
\newblock {\em arXiv preprint arXiv:2104.08691}, 2021.

\bibitem{li2021prefix}
Xiang~Lisa Li and Percy Liang.
\newblock Prefix-tuning: Optimizing continuous prompts for generation.
\newblock {\em arXiv preprint arXiv:2101.00190}, 2021.

\bibitem{liu2018darts}
Hanxiao Liu, Karen Simonyan, and Yiming Yang.
\newblock Darts: Differentiable architecture search.
\newblock {\em arXiv preprint arXiv:1806.09055}, 2018.

\bibitem{loshchilov2017decoupled}
Ilya Loshchilov and Frank Hutter.
\newblock Decoupled weight decay regularization.
\newblock {\em arXiv preprint arXiv:1711.05101}, 2017.

\bibitem{lu2022prompt}
Yuning Lu, Jianzhuang Liu, Yonggang Zhang, Yajing Liu, and Xinmei Tian.
\newblock Prompt distribution learning.
\newblock In {\em Proceedings of the IEEE/CVF Conference on Computer Vision and
  Pattern Recognition (CVPR)}, 2022.

\bibitem{maji2013fine}
Subhransu Maji, Esa Rahtu, Juho Kannala, Matthew Blaschko, and Andrea Vedaldi.
\newblock Fine-grained visual classification of aircraft.
\newblock {\em arXiv preprint arXiv:1306.5151}, 2013.

\bibitem{mao2021unipelt}
Yuning Mao, Lambert Mathias, Rui Hou, Amjad Almahairi, Hao Ma, Jiawei Han,
  Wen-tau Yih, and Madian Khabsa.
\newblock Unipelt: A unified framework for parameter-efficient language model
  tuning.
\newblock {\em arXiv preprint arXiv:2110.07577}, 2021.

\bibitem{matthey2017dsprites}
Loic Matthey, Irina Higgins, Demis Hassabis, and Alexander Lerchner.
\newblock dsprites: Disentanglement testing sprites dataset, 2017.

\bibitem{netzer2011reading}
Yuval Netzer, Tao Wang, Adam Coates, Alessandro Bissacco, Bo~Wu, and Andrew~Y
  Ng.
\newblock Reading digits in natural images with unsupervised feature learning.
\newblock 2011.

\bibitem{nilsback2006visual}
M-E Nilsback and Andrew Zisserman.
\newblock A visual vocabulary for flower classification.
\newblock In {\em Proceedings of the IEEE/CVF Conference on Computer Vision and
  Pattern Recognition (CVPR)}, volume~2, pages 1447--1454. IEEE, 2006.

\bibitem{parkhi2012cats}
Omkar~M Parkhi, Andrea Vedaldi, Andrew Zisserman, and CV~Jawahar.
\newblock Cats and dogs.
\newblock In {\em Proceedings of the IEEE/CVF Conference on Computer Vision and
  Pattern Recognition (CVPR)}, pages 3498--3505. IEEE, 2012.

\bibitem{pham2018efficient}
Hieu Pham, Melody Guan, Barret Zoph, Quoc Le, and Jeff Dean.
\newblock Efficient neural architecture search via parameters sharing.
\newblock In {\em International conference on machine learning (ICML)}, pages
  4095--4104. PMLR, 2018.

\bibitem{radford2021learning}
Alec Radford, Jong~Wook Kim, Chris Hallacy, Aditya Ramesh, Gabriel Goh,
  Sandhini Agarwal, Girish Sastry, Amanda Askell, Pamela Mishkin, Jack Clark,
  et~al.
\newblock Learning transferable visual models from natural language
  supervision.
\newblock In {\em International Conference on Machine Learning (ICML)}, pages
  8748--8763. PMLR, 2021.

\bibitem{real2017large}
Esteban Real, Sherry Moore, Andrew Selle, Saurabh Saxena, Yutaka~Leon Suematsu,
  Jie Tan, Quoc~V Le, and Alexey Kurakin.
\newblock Large-scale evolution of image classifiers.
\newblock In {\em International Conference on Machine Learning (ICML)}, pages
  2902--2911. PMLR, 2017.

\bibitem{recht2019imagenet}
Benjamin Recht, Rebecca Roelofs, Ludwig Schmidt, and Vaishaal Shankar.
\newblock Do imagenet classifiers generalize to imagenet?
\newblock In {\em International Conference on Machine Learning (ICML)}, pages
  5389--5400. PMLR, 2019.

\bibitem{reed2022generalist}
Scott Reed, Konrad Zolna, Emilio Parisotto, Sergio~Gomez Colmenarejo, Alexander
  Novikov, Gabriel Barth-Maron, Mai Gimenez, Yury Sulsky, Jackie Kay,
  Jost~Tobias Springenberg, et~al.
\newblock A generalist agent.
\newblock {\em arXiv preprint arXiv:2205.06175}, 2022.

\bibitem{vaswani2017attention}
Ashish Vaswani, Noam Shazeer, Niki Parmar, Jakob Uszkoreit, Llion Jones,
  Aidan~N Gomez, {\L}ukasz Kaiser, and Illia Polosukhin.
\newblock Attention is all you need.
\newblock {\em Advances in neural information processing systems (NeuIPS)}, 30,
  2017.

\bibitem{Veeling2018qh}
Bastiaan~S Veeling, Jasper Linmans, Jim Winkens, Taco Cohen, and Max Welling.
\newblock Rotation equivariant {CNNs} for digital pathology.
\newblock June 2018.

\bibitem{wang2019learning}
Haohan Wang, Songwei Ge, Zachary Lipton, and Eric~P Xing.
\newblock Learning robust global representations by penalizing local predictive
  power.
\newblock {\em Advances in Neural Information Processing Systems (NeuIPS)}, 32,
  2019.

\bibitem{xiao2010sun}
Jianxiong Xiao, James Hays, Krista~A Ehinger, Aude Oliva, and Antonio Torralba.
\newblock Sun database: Large-scale scene recognition from abbey to zoo.
\newblock In {\em 2010 IEEE computer society conference on computer vision and
  pattern recognition}, pages 3485--3492. IEEE, 2010.

\bibitem{yuan2021florence}
Lu~Yuan, Dongdong Chen, Yi-Ling Chen, Noel Codella, Xiyang Dai, Jianfeng Gao,
  Houdong Hu, Xuedong Huang, Boxin Li, Chunyuan Li, et~al.
\newblock Florence: A new foundation model for computer vision.
\newblock {\em arXiv preprint arXiv:2111.11432}, 2021.

\bibitem{zaken2021bitfit}
Elad~Ben Zaken, Shauli Ravfogel, and Yoav Goldberg.
\newblock Bitfit: Simple parameter-efficient fine-tuning for transformer-based
  masked language-models.
\newblock {\em arXiv preprint arXiv:2106.10199}, 2021.

\bibitem{zeiler2014visualizing}
Matthew~D Zeiler and Rob Fergus.
\newblock Visualizing and understanding convolutional networks.
\newblock In {\em European conference on computer vision (ECCV)}, pages
  818--833. Springer, 2014.

\bibitem{zhai2019large}
Xiaohua Zhai, Joan Puigcerver, Alexander Kolesnikov, Pierre Ruyssen, Carlos
  Riquelme, Mario Lucic, Josip Djolonga, Andre~Susano Pinto, Maxim Neumann,
  Alexey Dosovitskiy, et~al.
\newblock A large-scale study of representation learning with the visual task
  adaptation benchmark.
\newblock {\em arXiv preprint arXiv:1910.04867}, 2019.

\bibitem{zhang2022bamboo}
Yuanhan Zhang, Qinghong Sun, Yichun Zhou, Zexin He, Zhenfei Yin, Kun Wang,
  Lu~Sheng, Yu~Qiao, Jing Shao, and Ziwei Liu.
\newblock Bamboo: Building mega-scale vision dataset continually with
  human-machine synergy.
\newblock {\em arXiv preprint arXiv:2203.07845}, 2022.

\bibitem{zhong2021factual}
Zexuan Zhong, Dan Friedman, and Danqi Chen.
\newblock Factual probing is [mask]: Learning vs. learning to recall.
\newblock {\em arXiv preprint arXiv:2104.05240}, 2021.

\bibitem{zhou2021domain}
Kaiyang Zhou, Ziwei Liu, Yu~Qiao, Tao Xiang, and Chen~Change Loy.
\newblock Domain generalization: A survey.
\newblock {\em arXiv preprint arXiv:2103.02503}, 2021.

\bibitem{zhou2021learning}
Kaiyang Zhou, Jingkang Yang, Chen~Change Loy, and Ziwei Liu.
\newblock Learning to prompt for vision-language models.
\newblock {\em arXiv preprint arXiv:2109.01134}, 2021.

\bibitem{zhou2022conditional}
Kaiyang Zhou, Jingkang Yang, Chen~Change Loy, and Ziwei Liu.
\newblock Conditional prompt learning for vision-language models.
\newblock In {\em Proceedings of the IEEE/CVF Conference on Computer Vision and
  Pattern Recognition (CVPR)}, 2022.

\bibitem{zhou2021osnet}
Kaiyang Zhou, Yongxin Yang, Andrea Cavallaro, and Tao Xiang.
\newblock Learning generalisable omni-scale representations for person
  re-identification.
\newblock {\em IEEE Transactions on Pattern Analysis and Machine Intelligence
  (TPAMI)}, 2021.

\bibitem{zoph2016neural}
Barret Zoph and Quoc~V Le.
\newblock Neural architecture search with reinforcement learning.
\newblock {\em arXiv preprint arXiv:1611.01578}, 2016.

\bibitem{zoph2018learning}
Barret Zoph, Vijay Vasudevan, Jonathon Shlens, and Quoc~V Le.
\newblock Learning transferable architectures for scalable image recognition.
\newblock In {\em Proceedings of the IEEE/CVF Conference on Computer Vision and
  Pattern Recognition (CVPR)}, pages 8697--8710, 2018.

\end{thebibliography}
}



\end{document}